\documentclass[letterpaper, 10 pt, conference]{ieeeconf}  

\IEEEoverridecommandlockouts                              

\usepackage{graphics} 
\usepackage{amsmath} 
\usepackage{amssymb}  
\usepackage{cite}
\usepackage{graphicx}
\usepackage{subcaption}

\newcommand{\asimo}{\includegraphics[scale=0.06]{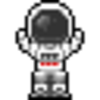}}

\title{\LARGE \bf
Neuro-Symbolic Imitation Learning: \\ Discovering Symbolic Abstractions for Skill Learning
}

\author{Leon Keller$^{1, \asimo, *}$, Daniel Tanneberg$^{\asimo}$, and Jan Peters$^{1, 2, 3}$
\thanks{*\texttt{mail@leon-keller.com}}
\thanks{$^{1}$ Intelligent Autonomous Systems, TU Darmstadt, Germany} %
\thanks{$^{2}$ German Research Center for AI, Germany}%
\thanks{$^{3}$ Hessian Centre for Artificial Intelligence, Germany}%
\thanks{\asimo Honda Research Institute EU, Germany}%
\thanks{Website: \href{https://hri-eu.github.io/NeuroSymbolicImitationLearning}{https://hri-eu.github.io/NeuroSymbolicImitationLearning}\vspace{-5pt}}%
}

\usepackage{tikz}
\usepackage{eso-pic}
\usepackage{hyperref}
\newcommand\copyrighttext{%
  \footnotesize \copyright 2025 IEEE. Personal use of this material is permitted. Permission from IEEE must be obtained for all other uses, in any current or future media, including reprinting/republishing this material for advertising or promotional purposes, creating new collective works, for resale or redistribution to servers or lists, or reuse of any copyrighted component of this work in other works.
  DOI: \href{https://doi.org/10.1109/ICRA55743.2025.11127692}{10.1109/ICRA55743.2025.11127692} 
}%
 
\newcommand\copyrightnotice[1][black]{%
\begin{tikzpicture}[remember picture,overlay]
\node[anchor=south,yshift=10pt,draw=#1] at (current page.south) {\parbox{\dimexpr\textwidth-\fboxsep-\fboxrule\relax}{\copyrighttext}};
\end{tikzpicture}%
}
 
\newcommand\overlaycopyrightnotice[1][black]{%
\AddToShipoutPicture*{\copyrightnotice[#1]}%
}

\begin{document}

\maketitle
\overlaycopyrightnotice
\thispagestyle{empty}
\pagestyle{empty}


\begin{abstract}
Imitation learning is a popular method for teaching robots new behaviors.
However, most existing methods focus on teaching short, isolated skills rather than long, multi-step tasks.
To bridge this gap, imitation learning algorithms must not only learn individual skills but also an abstract understanding of how to sequence these skills to perform extended tasks effectively.
This paper addresses this challenge by proposing a neuro-symbolic imitation learning framework.
Using task demonstrations, the system first learns a symbolic representation that abstracts the low-level state-action space. 
The learned representation decomposes a task into easier subtasks and allows the system to leverage symbolic planning to generate abstract plans. 
Subsequently, the system utilizes this task decomposition to learn a set of neural skills capable of refining abstract plans into actionable robot commands.
Experimental results in three simulated robotic environments demonstrate that, compared to baselines, our neuro-symbolic approach increases data efficiency, improves generalization capabilities, and facilitates interpretability.
\end{abstract}


\section{Introduction}
Over the last decade, imitation learning~\cite{imitation_zheng, imitation_fang, imitation_zare, osa, peters} has emerged as a powerful approach for teaching new behaviors to robots.
Although existing approaches excel at teaching isolated skills, real-world tasks often involve multiple steps that require combining a variety of skills. 
Thus, to effectively employ robots in our daily lives, we require algorithms capable of utilizing demonstrations to teach not only individual skills but also how these skills can be sequenced to solve extended, intricate tasks~\cite{manschitz2014learning,shiarlis2018taco,tanneberg2021skid}.
For instance, consider a robot designed to assist in the kitchen: To be a truly useful assistant, the robot must not only be able to execute individual skills -- such as placing a pot on the stove, adding water, and turning on the heat -- but also understand how to sequence these skills correctly to cook a complete meal.

Humans have a remarkable ability to tackle such long, complex tasks by relying on a fundamental cognitive tool: abstraction~\cite{tenenbaum2011grow,konidaris2019necessity,tanneberg2020evolutionary}.
We naturally simplify and distill information, recognize patterns, and abstract away noise and unnecessary information. 
This process not only aids in comprehending what we see and experience but also serves as the foundation for reasoning about our surroundings, planning complex behavior, and generalizing knowledge from one context to another.
Similarly, when programming robots to solve complex tasks, human engineers often rely on hierarchical approaches, one of the most popular being Task and Motion Planning (TAMP)~\cite{tamp_garrett, tamp_guo, tamp_mansouri}.
TAMP involves using a high-level symbolic abstraction of the state-action space to generate task plans, which are subsequently refined into actionable low-level commands through classical motion planning.
While this approach is effective, it requires considerable effort from a human expert:
Engineers typically must manually design symbolic representations for high-level planning and develop models for low-level motion planning.

\begin{figure*}[t] 
    \centering
    \includegraphics[width=0.98\textwidth]{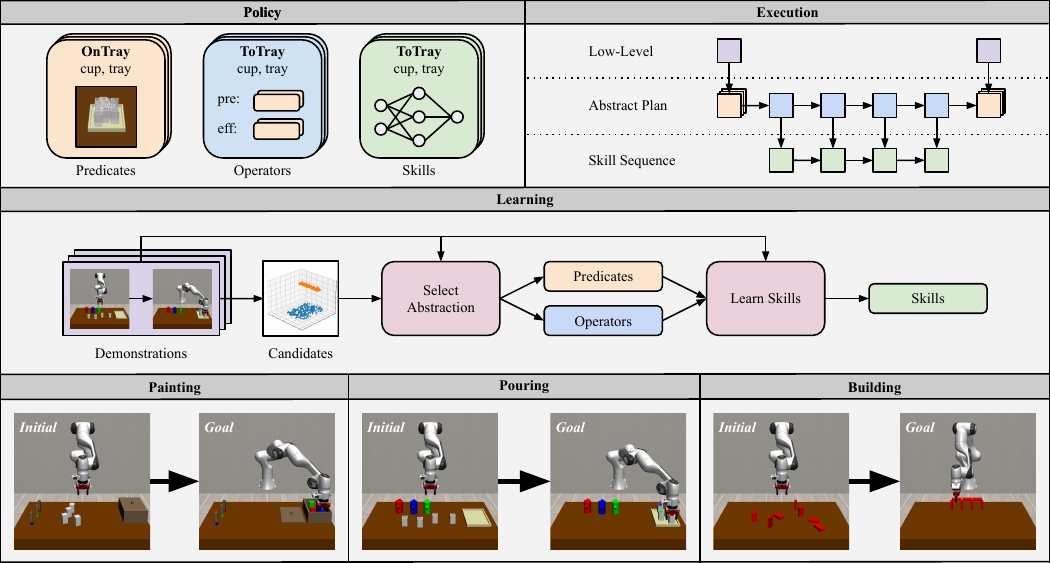}
    \caption{
    \textit{(Top Left)} The components of the neuro-symbolic policy. Predicates abstract the state-space, operators define an abstract transition model, and skills execute abstract plans. 
    \textit{(Top Right)} Illustration of the policy execution. First, the start and goal state are abstracted using the predicates. Following, an abstract plan is computed using the operators and planning algorithms. Lastly, the corresponding skill sequence is executed.
    \textit{(Middle)} Overview of the learning pipeline. First, a set of candidate abstractions is generated based on the demonstrations. Subsequently, a subset of these candidates is selected using a novel objective function. Lastly, the learned symbolic representation is utilized to learn a set of skills with behavior cloning.
    \textit{(Bottom)} The evaluation tasks. In each task, a panda robot has to manipulate objects placed on a table.}
    \label{fig:overview}
\end{figure*}

In this work, we propose a neuro-symbolic imitation learning framework that learns individual skills and planning capabilities from task demonstrations.
The neuro-symbolic policies, akin to TAMP, consist of symbolic components for abstract planning and neural components for refining abstract plans into actionable commands.
While previous work explored learning symbols when skills are given~\cite{Jetchev, bonet, rodriguez, konidaris, konidaris2, konidaris3, konidaris4, james, james2, ahmetoglu, ahmetoglu2, ahmetoglu3, asai, asai2, dittadi, umili, silver, loula, pasula, curtis, tanneberg2023learning}, and skills when symbols are given~\cite{acharya, guan, lyu, kokel, yang, hankui, illanes, grounds, cheng, silver2, mandlekar, mcdonald}, our approach aims at acquiring symbols \textit{and} skills from raw task demonstrations.
Our approach to learning neuro-symbolic policies is divided into two phases:
In the first phase, we learn the symbolic components of the policy, consisting of a set of predicates that abstract the state space and a set of operators that define a transition model in the abstract state space. 
Together, predicates and operators define a planning problem in the Planning Domain Definition Language~\cite{pddl} (PDDL).
To learn predicates, we first generate a set of candidate predicates based on features observed in the demonstrations. 
Subsequently, we select among these candidates by optimizing a novel objective function. 
Concurrently, operators are learned based on the symbolic transitions induced by the predicates.
In the second phase, we utilize the identified symbolic abstraction to learn a distinct neural skill for every operator identified in the first phase.
For that, we segment the demonstrations based on the learned symbolic representation and train the neural networks using standard behavior cloning.
Importantly, rather than training the neural skill on the full state, we constrain the input space to focus on the objects relevant to the current subtask, as defined by the symbolic representation. 
To execute a neuro-symbolic policy, we leverage off-the-shelf symbolic planning algorithms~\cite{fastdown, kstar, hasler2024efficient} to generate an abstract plan consisting of a sequence of operators. 
Subsequently, we utilize the corresponding neural skills to execute the resulting abstract plan on the robot. Fig.~\ref{fig:overview} shows an overview of the full approach.

We demonstrate the advantages of our neuro-symbolic framework by comparing it to baselines in three simulated robotic environments.
Experimental results show that our neuro-symbolic approach increases data efficiency, improves generalization capabilities, and facilitates interpretability.

\vspace*{-1cm}

\section{Related Work}
Our work is closely related to methods that learn symbolic representations that abstract the state-action space of decision-making problems, particularly those focused on acquiring abstractions that facilitate high-level task planning.

A long line of research focuses on learning symbols given a predefined set of skills:
Early work~\cite{Jetchev} defines predicates as radial basis function classifiers that are optimized to produce symbols that facilitate the learning of accurate transition and reward models for planning.
In~\cite{bonet, rodriguez}, a symbolic representation is learned from given labeled graphs that encode the structure of the state-space, where nodes correspond to states and edges to skills. The construction of the symbolic representation is encoded as a Boolean satisfiability~\cite{bonet} or answer set programming~\cite{rodriguez} problem. 
Another body of work learns a symbolic representation based on the initiation and termination sets of option policies in a semi-Markov decision process~\cite{konidaris, konidaris2, konidaris3, konidaris4, james, james2}.
Several works utilize neural networks with a binary bottleneck layer to learn symbols for planning~\cite{ahmetoglu, ahmetoglu2, ahmetoglu3, asai, asai2, dittadi, umili}. After training the network, the binary latent vector is used to create symbols for planning.
The learning objective of the networks is either reconstructing the state~\cite{asai, asai2, dittadi} or effect prediction~\cite{ahmetoglu, ahmetoglu2, ahmetoglu3, umili}.
Multiple works invent a symbolic representation leveraging a grammar or concept language~\cite{silver, loula, pasula, curtis}. The grammar is used to generate candidate abstraction, from which the most suitable are selected using differing objective functions.
The common objective of these methods is to learn symbols capable of generating high-level plans composed of predefined skills.
In contrast, our method acquires a symbolic representation directly from low-level demonstrations and subsequently utilizes this representation to \textit{learn} skills.
Closest to our work,~\cite{shah} learns relational symbols from raw demonstrations. 
Symbols are invented by identifying relational critical regions in the demonstrations.
However, this method does not consider skill learning but utilizes motion planning to refine abstract plans.

Moreover, our approach is related to works that leverage symbolic representations for skill learning. The integration of symbolic planning and reinforcement learning has been extensively explored in the literature~\cite{acharya, guan, lyu, kokel, yang, hankui, illanes, grounds, cheng}. These approaches use symbolic abstractions to guide the learning process and enhance generalization. Particularly relevant to our work are studies that utilize symbolic abstractions in imitation learning~\cite{silver2, chitnis, mandlekar, mcdonald}. Symbolic representations are used to break down tasks into simpler subtasks, enabling the learning of distinct policies for each subtask.

To the best of our knowledge, our neuro-symbolic framework is the first to simultaneously learn a relational symbolic abstraction capable of generating high-level plans and neural skills that refine these plans from raw demonstrations.


\section{Preliminaries}
In this work, we explore imitation learning in fully observable, robotic environments.
Each environment consists of a robot and multiple objects that the robot can manipulate.
Each object $o \in \mathcal{O}$ is associated with a type $t(o) \in \mathcal{T}$ and described by several feature vectors $\boldsymbol{\xi}^{i}(o) \in \Xi(o)$, which collectively represent the object's state $\boldsymbol{s}_{t}(o)$ in the world.
Different types of objects may be described by different feature vectors. 
For example, the state of an object $o_1$ of type $t(o_1) = \mathrm{eef}$ may contain its cartesian pose and whether the gripper is opened or closed, 
while the state of an object $o_2$ of type $t(o_2) = \mathrm{cube}$ may include the color of the cube.
We define the environments state $\boldsymbol{s}_{t} \in \mathcal{S}$ at time step $t$ as the concatenation of the states of all objects in the environment $\boldsymbol{s}_{t} = [\boldsymbol{s}_{t}(o_1), \ \boldsymbol{s}_{t}(o_2), \ \boldsymbol{s}_{t}(o_3), \ \ldots]$.
Consequently, the dimensionality of the state space $\mathcal{S}$ depends on the number of objects in the environment and their respective types.
The robot is controlled using an Operational Space Controller.
An action $\boldsymbol{a} \in \mathcal{A}$ specifies an offset to the current end-effector pose, producing the target pose for the controller.
Executing an action $\boldsymbol{a}_{t} \in \mathcal{A}$ transitions the system into a new state $\boldsymbol{s}_{t+1} \in \mathcal{S}$ following the environment's dynamics.
We define a trajectory 
$\boldsymbol{\tau} = (\boldsymbol{s}_0, \ \boldsymbol{a}_0, \ \ldots, \ \boldsymbol{s}_N)$ 
as a sequence of state-action pairs obtained by applying actions in the environment.
A task is represented by a set of goal states $\mathcal{S}_g \subset \mathcal{S}$ and a trajectory is considered to solve that task if $\boldsymbol{s}_N \in \mathcal{S}_g$.

During training, we assume access to a set of trajectories $\mathcal{D} = \{\boldsymbol{\tau}_{0}, \ \ldots, \ \boldsymbol{\tau}_{M}\}$ that demonstrate how to successfully solve a set of tasks in the environment.
The objective is to learn a neuro-symbolic policy capable of solving tasks that differ from those encountered in the demonstrations. 
These generalization variations include unseen initial and goal object configurations, and tasks involving a greater number of objects than seen during training.


\section{Neuro-Symbolic Policies} 
In our framework, policies have both symbolic and neural components. 
The symbolic components consist of predicates $\mathcal{P}$ that abstract the state-space and operators $\Sigma$ that define a transition model in the abstract state-space induced by the predicates. 
Together, predicates and operators define a planning problem in the Planning Domain Definition Language (PDDL)~\cite{pddl} and can be utilized to generate abstract plans.
The neural components consist of skills $\Pi$ that together enable the execution of abstract plans in the environment. 

\subsubsection{Predicates} 
A predicate $p \in \mathcal{P}$ is a binary function with typed parameters $\Theta$ that specifies a relation between a number of objects. 
For instance, the predicate $\mathrm{onTop}(\mathrm{cube}, \mathrm{cube})$ specifies the $\mathrm{onTop}$ relation between objects of type $\mathrm{cube}$.
A grounded predicate $\hat{p}$ is the truth value of a predicate when evaluated for specific objects, e.g., for given objects $o_1, \ o_2$ with $t(o_1) = t(o_2) = \mathrm{cube}$ the grounded predicate $\mathrm{onTop}(\mathrm{o_1}, \mathrm{o_2}) = \mathrm{True}$ specifies that $o_1$ is on top of $o_2$.
Given an environments state $\boldsymbol{s} \in \mathcal{S}$ and a set of predicates $\mathcal{P}$, we define the corresponding abstract state $\bar{\boldsymbol{s}} = \psi(\boldsymbol{s}, \mathcal{P})$ as the set of all grounded predicates that are $\mathrm{True}$ in $\boldsymbol{s}$.
Likewise, we define an abstract trajectory $\bar{\boldsymbol{\tau}} = \psi(\boldsymbol{\tau}, \mathcal{P})$ as the trajectory obtained when abstracting every state in $\boldsymbol{\tau}$. 
Grounded predicates typically do not change their truth value at every time step and, thus, $|\bar{\boldsymbol{\tau}}|$ \(\ll\) $|\boldsymbol{\tau}|$.

\subsubsection{Operators} 
An operator $\sigma \in \Sigma$ is characterized by a set of typed parameters $\Theta$ which specify the types of objects the operator can act on, positive and negative precondition sets $\mathrm{pre}_{+}, \ \mathrm{pre}_{-} \subset \mathcal{P}$, as well as positive and negative effect sets $\mathrm{eff}_{+}, \ \mathrm{eff}_{-} \subset \mathcal{P}$.
A grounded operator $\hat{\sigma}$ is a binding of specific objects to the typed parameters of an operator. This process grounds all the predicates in the operator's precondition and effect sets.
A grounded operator is applicable in
$\bar{\boldsymbol{s}}$, if $\forall \hat{p} \in \mathrm{pre}_{+}: \hat{p} \in \bar{\boldsymbol{s}} \ \land \ \forall \hat{p} \in \mathrm{pre}_{-}: \hat{p} \notin \bar{\boldsymbol{s}}$.
Its execution generates the next abstract state by changing all grounded predicates in $\bar{\boldsymbol{s}}$ according to its effect sets.

\subsubsection{Skills} 
The responsibility of a skill $\pi^i \in \Pi$ is to execute the operator it is linked to by interacting with the environment. 
Specifically, when a skill $\pi^{i}$ is executed in a state $s \in \mathcal{S}$, where the corresponding operator $\sigma^{i}$ is applicable, the skill's role is to transition the environment into a state in which the operator’s effects are satisfied.
As the symbolic representation abstracts the state space, many different states $s \in \mathcal{S}$ can satisfy the effects of a given operator. 
For example, consider an operator whose effect is defined by a predicate that measures whether a cup is placed on a tray.
The corresponding skill should not be restricted to placing the cup at a single fixed position on the tray. 
Instead, the skill should be able to place the cup at multiple different positions, depending on the concrete circumstances.
Thus, a skill $\pi^i = (f^i, g^i) \in \Pi$ in our framework consists of two components: 
a subgoal sampler $g^i$ and a subgoal-conditioned controller $f^i$. 
The subgoal sampler $g^i$ is responsible for proposing environment states that fulfill the operator's effects, while the subgoal-conditioned controller $f^i$ is responsible for transitioning the environment toward the subgoal proposed by the sampler $g^i$ by iteratively executing actions $a \in \mathcal{A}$ in the environment.

\subsection{Executing a Neuro-Symbolic Policy}
To solve a task with the neuro-symbolic policy, we first generate an abstract plan using the predicates $\mathcal{P}$ and operators $\Sigma$, and then execute the abstract plan in the environment using the skills $\Pi$.

\subsubsection{Generate Abstract Plan}
Given an initial state $\boldsymbol{s}_0 \in \mathcal{S}$ and a set of goal states $\mathcal{S}_g \subset \mathcal{S}$, we first compute an abstract initial and goal state.
The abstract initial state is given by abstracting $\boldsymbol{s}_0$ using the predicates $\bar{\boldsymbol{s}}_0 = \psi(\boldsymbol{s}_0, \mathcal{P})$.
Moreover, the abstract goal state $\bar{\boldsymbol{s}}_g$ is given by the set of all grounded predicates that hold true across all $\boldsymbol{s}_g \in \mathcal{S}_g$.
Next, we seek a sequence of grounded operators $(\hat{\sigma}^i_0, \hat{\sigma}^j_1, \ldots)$ that induce a symbolic trajectory from $\bar{\boldsymbol{s}}_0$ to $\bar{\boldsymbol{s}}_g$, referred to as an abstract plan.
Since the symbolic components are represented in PDDL, we can employ off-the-shelf symbolic planning algorithms~\cite{fastdown, kstar, hasler2024efficient} to generate abstract plans. 
Specifically, we use a top-k planner to generate the best $K$ abstract plans, assuming unit cost for each operator.
Since the symbolic components abstract the environment's state and action space, not each of these abstract plans is guaranteed to be downwards refineable~\cite{downward_refine, tamp_garrett}. 
Thus, among the generated plans, we select the one most similar to the abstract plans observed in the demonstrations. To do this, we compute the minimum Levenshtein distance~\cite{levenshtein1, levenshtein2} between each generated abstract plan and the demonstrated plans, choosing the one with the smallest minimum distance.

\subsubsection{Execute Abstract Plan}
Once an abstract plan is selected, the next step is to sequentially execute the corresponding skill for each grounded operator in the plan. 
The execution of a skill $\pi^i$ involves two steps: 
First, the skill's subgoal sampler proposes a subgoal consistent with the effects of the corresponding grounded operator $\hat{\sigma}^i$. 
Following, the skill's subgoal-conditioned controller is used to transition the environment towards the desired environment state. 
The controller continues executing actions until the effects of the corresponding grounded operator are satisfied. 
Once the operator's effects are achieved, the process moves on to the next grounded operator and corresponding skill in the plan.


\begin{figure*}[t]
    \centering
    \begin{minipage}{0.54\textwidth} 
        \centering
        \includegraphics[width=1.23in]{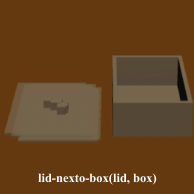}
        \includegraphics[width=1.23in]{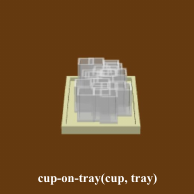}
        \includegraphics[width=1.23in]{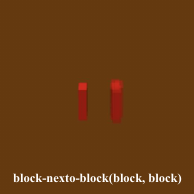}
        \caption{Visualization of learned predicates. Predicates are visualized by overlaying images of states in which the predicate is true.}
       \label{fig:predicate_examples}
    \end{minipage}
    ~
    \begin{minipage}{0.36\textwidth} 
        \centering
        \includegraphics[width=1.23in]{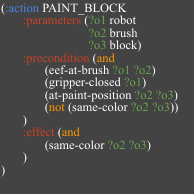}
        \includegraphics[width=1.23in]{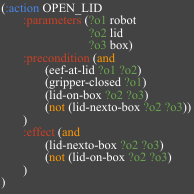}
        \caption{Illustration of learned operators. The operators are shown in PDDL-Syntax.}
        \label{fig:operator_examples}
    \end{minipage}
\end{figure*}

\section{Learning Neuro-Symbolic Policies}
Our main contribution is twofold: 
First, we propose a novel method for learning predicates from raw task demonstrations. 
Second, to the best of our knowledge, our approach is the first to combine predicate, operator, and skill learning into a single, unified framework.

\subsection{Skill Learning}
\label{sec:skill_learning_full}
While in our framework, the symbolic representation is learned first and then used to acquire skills, here we first discuss skill learning to enhance clarity.
Our approach to skill learning is similar to previous work~\cite{silver2}. 
Given predicates $\mathcal{P}$ and operators $\Sigma$, we learn a subgoal-conditioned controller $f_i$ and a subgoal sampler $g_i$ for each operator $\sigma_i \in \Sigma$. 
In contrast, we transform the states differently and adopt non-parametric distributions to represent the samplers.

\subsubsection{Generate Skill Datasets}
\label{sec:skill_dataset}
We decompose the demonstrations into distinct skill datasets by segmenting each $\boldsymbol{\tau} \in \mathcal{D}$ according to the abstraction defined by the predicates $\mathcal{P}$ and operators $\Sigma$.
A segment $(j, k)$ is induced by each sub-sequence $(\boldsymbol{s}_j, \boldsymbol{a}_j, \ldots, \boldsymbol{s}_k, \boldsymbol{a}_k)$ of a trajectory, during which the abstract state remains constant. We define the state that immediately follows the segment $\boldsymbol{s}_{k+1}$ as the segments' goal.
We determine to which skill dataset each segment $(i, j)$ should be added by examining which operator's $\sigma$ preconditions and effects match to the transition from the abstract state of the segment $\psi(\boldsymbol{s}_{(j:k)}, \mathcal{P})$ to the abstract state of the segment's goal $\psi(\boldsymbol{s}_{k+1}, \mathcal{P})$.
Before adding the segment to the skill dataset, we transform the states to only contain information relevant to the execution of operator $\sigma$.
For that, we define a transformation function $\phi_{\sigma}$, which maps a state $\boldsymbol{s}$ to a transformed state based on the predicates in the effect sets of the operator. 
For instance, if the effect sets only contain the predicate $\mathrm{OnTop}(o_1, o_2) = \mathrm{True}$, then $\phi_{\sigma}(\boldsymbol{s})$ will only return the relative pose between $o_1$ and $o_2$.
We then add the transformed segment to the skill dataset corresponding to operator $\sigma$: $D_{\sigma} = D_{\sigma} \cup \{(\phi_{\sigma}(\boldsymbol{s}_l), \ \boldsymbol{a}_l, \ \phi_{\sigma}(\boldsymbol{s}_{k+1}))\}_{l = j:k}$.

\subsubsection{Learn Skills}
\label{sec:skill_learning} 
After the datasets are constructed, we learn a skill $\pi^i = (f^i, g^i)$ for each dataset $D_{\sigma^i}$.
Learning a goal-conditioned controller $f^i$ reduces to a supervised learning problem. 
For each $f^i$, we train a multi-layer perception using behavior cloning.
Learning a goal sampler $g^i$ is framed as a probability density estimation problem. 
For each $g^i,$ we model the distribution of possible goals using kernel density estimation with radial basis function kernels. 

\subsection{Operator Learning}
\label{sec:op_learning}
Next, as our method utilizes operator learning for scoring candidate predicates, we detail how operators $\Sigma$ are learned for a given set of predicates $\mathcal{P}$.
In the remainder of this work, we denote the operators induced by predicates $\mathcal{P}$ and demonstrations $\mathcal{D}$ as $\Sigma(\mathcal{P}, \mathcal{D})$.

To learn operators, we leverage symbolic state transitions observed in the demonstrations $\mathcal{D}$.
Specifically, for each pair of consecutive symbolic states $\bar{\boldsymbol{s}}_t, \bar{\boldsymbol{s}}_{t+1} \in \psi(\boldsymbol{\tau}, \mathcal{P}), \boldsymbol{\tau} \in \mathcal{D}$, we construct a tuple $z = (\mathrm{pre}_{+}, \mathrm{pre}_{-}, \mathrm{eff}_{+}, \mathrm{eff}_{-})$.
In this tuple, $\mathrm{pre}_{+}$ contains all the grounded predicates that are true in $\bar{\boldsymbol{s}}_t$, while $\mathrm{pre}_{-}$ contains those that are false.
Similarly, $\mathrm{eff}_{+}$ contains all the grounded predicates that become true when transitioning from $\bar{\boldsymbol{s}}_t$ to $\bar{\boldsymbol{s}}_{t+1}$, while $\mathrm{eff}_{-}$ contains those that become false. 
We then introduce a typed parameter $\theta$ for each object involved in the effect sets of the tuple and replace the objects in all the grounded predicates with the corresponding typed parameter.
Objects that are not part of the effect sets remain unchanged and are treated as constants.
Thus, after introducing these parameters, each tuple consists of a set of typed parameters $\Theta$ and lifted precondition and effect sets.

Following, we group the tuples $z \in Z$ based on their lifted effect sets. 
Specifically, two tuples $z_i, z_j \in Z$ belong to the same group $G_k$ if and only if their effect sets and typed parameters are identical.
We then construct an operator $\sigma_k$ for every group $G_k$: 
The operators typed parameters $\Theta(\sigma_k)$ and effect sets $\mathrm{eff}_{+}(\sigma_k), \ \mathrm{eff}_{-}(\sigma_k)$ are defined by the groups typed parameters and effect sets. 
The operator's precondition sets are given by the intersection of the corresponding precondition sets of all tuples in the group: $\mathrm{pre}_{+}(\sigma_k) = \cap_{z_i \in G_{k}} \mathrm{pre}_{+}(z_i), \ \mathrm{pre}_{-}(\sigma_k) = \cap_{z_i \in G_{k}} \mathrm{pre}_{-}(z_i)$.
This assumes that for every lifted effect set in the demonstrations, there is exactly one lifted precondition set. 
Similar approaches for operator learning are widely used in the literature~\cite{wang, stern, silver, silver2, shah, chitnis, tanneberg2023learning, verma}.

\subsection{Predicate Learning}
\label{sec:pred_learning}
We learn predicates in a two-stage process. First, we generate a set of candidate predicates $\mathcal{C}$ by clustering absolute and relative features observed in the demonstrations.
Then, we select a subset $\mathcal{P} \subset \mathcal{C}$ from these candidate predicates that trades off between a fine-grained segmentation and the complexity of the induced operator set.

\subsubsection{Generate Candidates}
\label{sec:generate_candidates}
Since the predicates should capture relations between objects, we generate candidate predicates based on relative features observed in the demonstrations. 
For each pair of object types $t_1, t_2 \in \mathcal{T}$ and each shared feature $\xi$ between them, we construct a relative feature dataset $D_{t_1, t_2}^{\xi}$.
For example, consider the object types $t_1 = \mathrm{cup}$ and $t_2 = \mathrm{tray}$, and the feature $\xi = \mathrm{pos}$ representing the objects' cartesian positions. 
To populate the relative feature dataset $D_{\mathrm{cup}, \mathrm{tray}}^{\mathrm{pos}}$, we iterate over all environment states $s_t \in \tau, \tau \in \mathcal{D}$, and compute the relative position $\Delta \mathrm{pos}_{t}(o_1, o_2) = \mathrm{pos}_{t}(o_1) - \mathrm{pos}_{t}(o_2)$ for all object pairs $o_1 \in O(\mathrm{cup}), \ o_2 \in O(\mathrm{tray})$.
For non-position-based features, other appropriate difference functions are used. For example, to populate relative orientation datasets, we compute relative quaternions.
Each relative feature is added to the dataset only if it remains constant between $s_t$ and the consecutive state $s_{t+1}$. 
This ensures that the predicates capture stationary relations rather than transient ones during movement.

Once a relative feature dataset $D_{t_1, t_2}^{\xi}$ is constructed, we apply agglomerative clustering to identify dense regions in the feature space. 
Initially, each relative feature in the dataset is treated as its own cluster. 
We then iteratively merge the two closest clusters until no two clusters are closer than the hyperparameter $\epsilon$.
Fig.~\ref{fig:overview} show the final clusters for the relative feature dataset $D_{\mathrm{cup}, \mathrm{tray}}^{\mathrm{pos}}$.
Following, a candidate predicate is created for each cluster. The typed parameters of the candidate predicate are defined as $\Theta = (t_1, t_2)$. 
For objects $o_1 \in O(t_1), \ o_2 \in O(t_2)$ and environment state $s_t \in \mathcal{S}$, the grounded instance of the predicate evaluates to $\mathrm{True}$ if the minimum distance between $\Delta \mathrm{\xi}_{t}(o_1, o_2)$ and the cluster is smaller than the threshold $\epsilon$.
Following the same procedure, we generate unary candidate predicates by constructing and clustering datasets that contain absolute feature values observed in the demonstrations.

\subsubsection{Select Abstraction}
\label{sec:select_abstraction}
Next, we select a subset $\mathcal{P} \subset \mathcal{C}$ of these candidates as the final predicates set $\mathcal{P}$ and learn the corresponding operators $\Sigma(\mathcal{P}, \mathcal{D})$.

Our primary objective is to learn a symbolic representation that facilitates efficient skill learning.
A predicate set segments a trajectory into subtasks by inducing a segment for each subsequence of a trajectory where the abstract state remains constant.
This segmentation is critical because it defines the skills that will be learned:
The shorter the segments are, the less complex the behavior is that the corresponding skill needs to learn.
Therefore, we want the segmentation to be as fine-grained as possible, meaning that we aim to maximize the number of abstract states induced by the predicates.
However, simply maximizing $\sum_{\tau \in \mathcal{D}} |\psi(\mathcal{P}, \tau)|$ will lead to selecting predicates that are overly specific to individual demonstrations, which would result in a large number of operators.
This can be detrimental to the generalization capabilities of the symbolic planning domain, as overly specific operators are less likely to generalize to new tasks.
Moreover, a large number of operators would also require a large number of skills. 
This is detrimental to skill learning, as a large number of skills means that each skill has access to less demonstration data. 
Thus, we regularize the objective by subtracting the number of operators $|\Sigma(\mathcal{P}, \mathcal{D})|$ induced by the predicate set.

Furthermore, the learned symbolic representation should generate sensible abstract plans.
We assume that the abstract plans in the demonstrations are optimal.
Thus, plans generated within the symbolic representation should not be shorter than the corresponding symbolic plans extracted from the demonstrations. 
If they were, they would be not downwards refineable and would likely miss an important predicate.
Therefore, we add a constraint that ensures that the abstract plans in the demonstrations are optimal plans in the symbolic representation.
Together, we get the optimization problem
\begin{align} 
\label{eq:opti}
&\max_{\mathcal{P} \subset \mathcal{C}} \sum_{\boldsymbol{\tau} \in \mathcal{D}} | \psi(\mathcal{P}, \boldsymbol{\tau}) | - \alpha | \Sigma(\mathcal{P}, \mathcal{D}) | \\
\mathrm{s.t.} \ | &\psi(\mathcal{P}, \boldsymbol{\tau}) | = |\mathrm{plan}(\mathcal{P}, \Sigma, \boldsymbol{\tau}_0, \boldsymbol{\tau}_N)| \ \forall \boldsymbol{\tau} \in \mathcal{D} \quad , \nonumber
\end{align}
where $\alpha$ controls the trade-off between a fine-grained segmentation and the complexity of the induced operator set.

We optimize (\ref{eq:opti}) with a beam search:
Initially, the beam contains only the empty predicate set.
In every iteration, we evaluate all predicate sets that can be created by adding a predicate $p \in C$ to one of the predicate sets in the beam and keep only the top $B$ subsets with the highest scores, where $B$ is the beam width.
This iterative process continues until no further improvement can be made.
While this beam search does not guarantee an optimal solution, we find that it consistently produces strong results for our objective.
Lastly, we select the highest scoring predicate set $\mathcal{P}$ and corresponding operator set $\Sigma(\mathcal{P}, \mathcal{D})$ of the beam that fulfills the constraint.


\section{Experiments}

\begin{figure*}[t]
    \centering
    \includegraphics[width=0.98\textwidth,trim={0 2.3mm 0 2.6mm},clip]{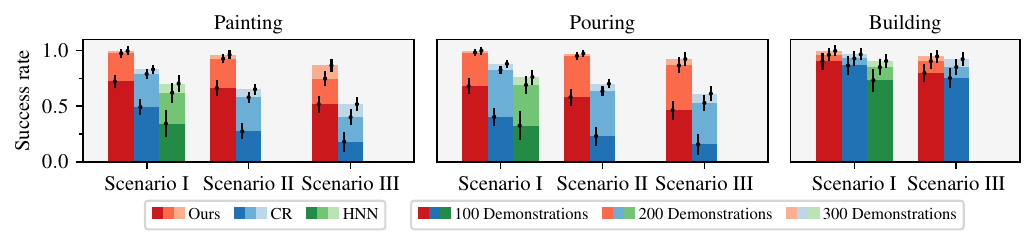}
    \caption{Comparison between our approach and baselines in three robotic environments.
    The x-axis denotes the different generalization scenarios; the y-axis denotes the success rate. Results are averaged over 10 random seeds, with bars representing the mean success rate and black lines indicating the standard deviation. Color shades denote the number of demonstrations used during training.}
    \label{fig:results}
\end{figure*}

We evaluate our method across three simulated robotic environments.
All environments utilize the \texttt{MuJoCo}~\cite{mujoco} physics engine and the \texttt{robosuite}~\cite{robosuite} simulation framework.
In each environment, a panda robot is positioned in front of a table with various objects placed on it. 
In the \texttt{Building} environment, the robot must assemble rectangular blocks into a bridge-like structure. 
For that, the robot must pick and place blocks in the correct order to prevent the bridge from collapsing.
In the \texttt{Pouring} environment, the robot must fill cups with tea and subsequently place the filled cups on a tray. 
There are three teapots, and each contains a different type of tea. 
In the \texttt{Painting} environment, the robot must paint blocks using a brush and subsequently place the painted blocks into a box.
The lid of the box needs to be opened before cubes can be placed in it.
There are three different brushes, each capable of applying a different color. 
The bottom row of Fig.~\ref{fig:overview} shows an exemplary initial and goal state for each environment.

We compare our approach to two baselines: 
\textit{Critical Region (CR)} ablates the proposed objective function. 
Instead of optimizing the proposed objective, we score and select predicates based on the notion of criticality as defined in~\cite{shah}. 
\textit{Hierarchical Neural Network (HNN)} ablates symbolic planning by replacing it with a neural high-level policy. Specifically, we train a neural network that takes the current and goal state as input and predicts which skill to execute.

\subsection{Results}
We conduct experiments to evaluate the following questions: \textit{(Q1)} Can our approach learn neuro-symbolic policies data-efficiently? \textit{(Q2)} Do the learned neuro-symbolic policies generalize to tasks with unseen goals and unseen number of objects? \textit{(Q3)} Are the learned symbols interpretable?

We evaluate all approaches across three distinct generalization scenarios: \textit{Scenario I} introduces initial object poses not seen during training. 
\textit{Scenario II} additionally introduces unseen goals.
These unseen goals include new color combinations for the \texttt{Painting} environment and new tea combinations for the \texttt{Pouring} environment.
Lastly, \textit{Scenario III} introduces more objects than during training.

Fig.~\ref{fig:results} shows the average success rates across differing numbers of demonstrations.
With 300 training demonstrations, our method achieves a high success rate across all environments and generalization scenarios. Furthermore, it outperforms both baselines across all number of demonstrations, showcasing its data-efficiency \textit{(Q1)}.
While \textit{(HNN)} fails to generalize to tasks in \textit{Scenario II} and \textit{Scenario III}, our approach consistently maintains a high success rate \textit{(Q2)}.
This result highlights a major advantage of the neuro-symbolic approach: Through the learned symbols, we can benefit from the generalization capabilities of symbolic planning.

The second baseline \textit{(CR)} learns more complex symbolic representations than our approach across all three tasks. It learns more predicates and operators, and the learned operators, on average, have more parameters.
This increased complexity leads to poorer generalization of the symbolic representation to tasks in \textit{Scenario II} and \textit{Scenario III}. 
Furthermore, the larger number of operators requires a corresponding increase in the number of skills, and the additional parameters per operator complicate the state space that each skill must handle.
These factors collectively contribute to \textit{(CR)}'s lower performance relative to our method.

To address \textit{(Q3)}, we visualize each learned predicate by overlaying images of states in which the predicate is true. 
These visualizations allow us to assign meaningful names to all predicates, making them easier to interpret. 
Fig.~\ref{fig:predicate_examples} showcases the visualization of three learned predicates.
Once predicates are named, we can interpret the preconditions and effects of each operator and assign meaningful names to them. 
Fig.~\ref{fig:operator_examples} illustrates two learned operators.
With all symbols named, the abstract plans generated by the policy become fully interpretable.


\section{Conclusions}
In this work, we introduced a neuro-symbolic imitation learning framework that learns a symbolic planning domain and neural skills from task demonstrations.
The conducted experiments show that, compared to baselines, the resulting neuro-symbolic policies offer greater data efficiency, improved generalization, and better interpretability.

For future work, we plan to validate the framework on real robots to further validate its practical applicability. Additionally, we hypothesize that the neuro-symbolic approach offers even more benefits in multi-task settings and, consequently, aim to apply it in multi-task imitation learning.
Lastly, we aim to explore online adaptation of both the skills and the symbolic representation to enable generalization to new object categories and failure recovery.







\bibliographystyle{IEEEtran}
\bibliography{root}

\end{document}